\documentclass[journal]{IEEEtran}
\usepackage{graphicx}
\usepackage{subcaption}
\usepackage{multirow}
\usepackage[table]{xcolor}
\usepackage{xcolor}
\usepackage{amssymb}
\usepackage{color}
\usepackage{colortbl}
\usepackage{tabularx}
\usepackage{bm}
\usepackage{booktabs}
\usepackage{url}
\usepackage{stackengine}
\usepackage{float}
\usepackage{verbatim}
\usepackage{array}
\usepackage[colorlinks,linkcolor=blue]{hyperref}

\ifCLASSINFOpdf
\else
\fi

\begin{document}
\title{MP-ResNet: Multi-path Residual Network for\\the Semantic segmentation of\\ High-Resolution PolSAR Images}

\author{Lei~Ding, Kai~Zheng, Dong~Lin, Yuxing~Chen, Bing~Liu, Jiansheng~Li and Lorenzo~Bruzzone,~\IEEEmembership{Fellow,~IEEE,}
\thanks{L. Ding, Y. Chen and L. Bruzzone are with the Department
of Information Engineering and Computer Science, University of Trento,
38123 Trento, Italy (E-mail: lei.ding@unitn.it, hao.tang@unitn.it, lorenzo.bruzzone@unitn.it .}
\thanks{K. Zheng, B. Liu and J. Li are with the Information Engineering University, 450001 Zhengzhou, China.}
\thanks{D. Lin is with the Space Engineering University NCO School, 102200 Beijing, China.}
\thanks{Manuscript under review.}}

\markboth{IEEE GEOSCIENCE AND REMOTE SENSING LETTERS}%
{Shell \MakeLowercase{\textit{et al.}}: Bare Demo of IEEEtran.cls for Journals}

\maketitle

\begin{abstract}
There are limited studies on the semantic segmentation of high-resolution Polarimetric Synthetic Aperture Radar (PolSAR) images due to the scarcity of training data and the complexity of managing speckle noise. The Gaofen contest has provided open access a high-quality PolSAR semantic segmentation dataset. Taking this opportunity, we propose a Multi-path ResNet (MP-ResNet) architecture for the semantic segmentation of high-resolution PolSAR images. Compared to conventional U-shape encoder-decoder convolutional neural network (CNN) architectures, the MP-ResNet learns semantic context with its parallel multi-scale branches, which greatly enlarges its valid receptive fields and improves the embedding of local discriminative features. In addition, MP-ResNet adopts a multi-level feature fusion design in its decoder to effectively exploit the features learned from its different branches. Ablation studies show that the MP-ResNet has significant advantages over its baseline method (FCN with ResNet34). It also surpasses several classic state-of-the-art methods in terms of overall accuracy (OA), mean F1 and fwIoU, with only a limited increase of computational costs. This CNN architecture can be used as a baseline method for future studies on the semantic segmentation of PolSAR images. The code is available at: \href{https://github.com/ggsDing/SARSeg}{\textit{https://github.com/ggsDing/SARSeg}}.
\end{abstract}


%
\IEEEpeerreviewmaketitle

\section{Introduction}

\IEEEPARstart{S}{ynthetic} Aperture Radar (SAR) has been widely used in Earth observation applications due to its capability to work under all weather and daylight conditions. The semantic segmentation of Polarimetric SAR (PolSAR) images, namely the pixel-wise classification of PolSAR images according to ground surface types, is beneficial to a large number of remote sensing applications (e.g., urban area management, disaster monitoring and land-cover mapping).

In recent years, with the rise of convolutional neural networks (CNNs) many methods have been developed for the semantic segmentation of natural images \cite{zhao2017pspnet}\cite{chen2018deeplabv3+} and remote sensing images \cite{ding2020lanet}\cite{mou2019relation}. However, a limited number of studies has been conducted on the semantic segmentation of PolSAR images based on deep CNNs \cite{geng2020multi}. There are two major barriers: i) PolSAR images contain intense speckle noise due to the coherent imaging mechanism of PolSAR systems. This speckle noise is a severe challenge for the automatic segmentation algorithms. ii) Large amount of training data is required to train an effective deep CNN. The manual annotation of SAR data is not only labor-intensive but also difficult, since the ground objects in SAR images can hardly be recognized by human observation without assisting data and expert knowledge \cite{duan2018multi}.

Most of the literature works on the semantic segmentation of SAR images are designed under the assumption that there are not enough labeled samples to train a CNN. In \cite{li2018novel}, a relatively small portion of pixels of the image to be classified are used for the training phase. The Fully Connected Network (FCN) has been used in a pipeline that contains sliding-crop and sparse coding operations. In \cite{wang2018hierarchical}, the FCN is combined with the sparse and low-rank subspace representations to alleviate the problem of insufficient training data. In \cite{duan2018multi}, a multi-scale CNN has been proposed for the semantic segmentation of SAR images. To solve the problem of lacking ground truth segmentation maps, it uses image scene labels for the training. In \cite{cao2019pixel}, the FCN is used in the complex domain of SAR data to include the phase information. In \cite{wu2019polsar}, extra datasets are used to pre-train CNN models for the semantic segmentation of SAR images. Since all these studies have been conducted on small datasets (most datasets contain only a single image), their objectives are mainly to reduce the dependence on training data keeping high generalization capabilities.

\begin{figure*}[htbp]
    \centering
    \includegraphics[width=1\linewidth]{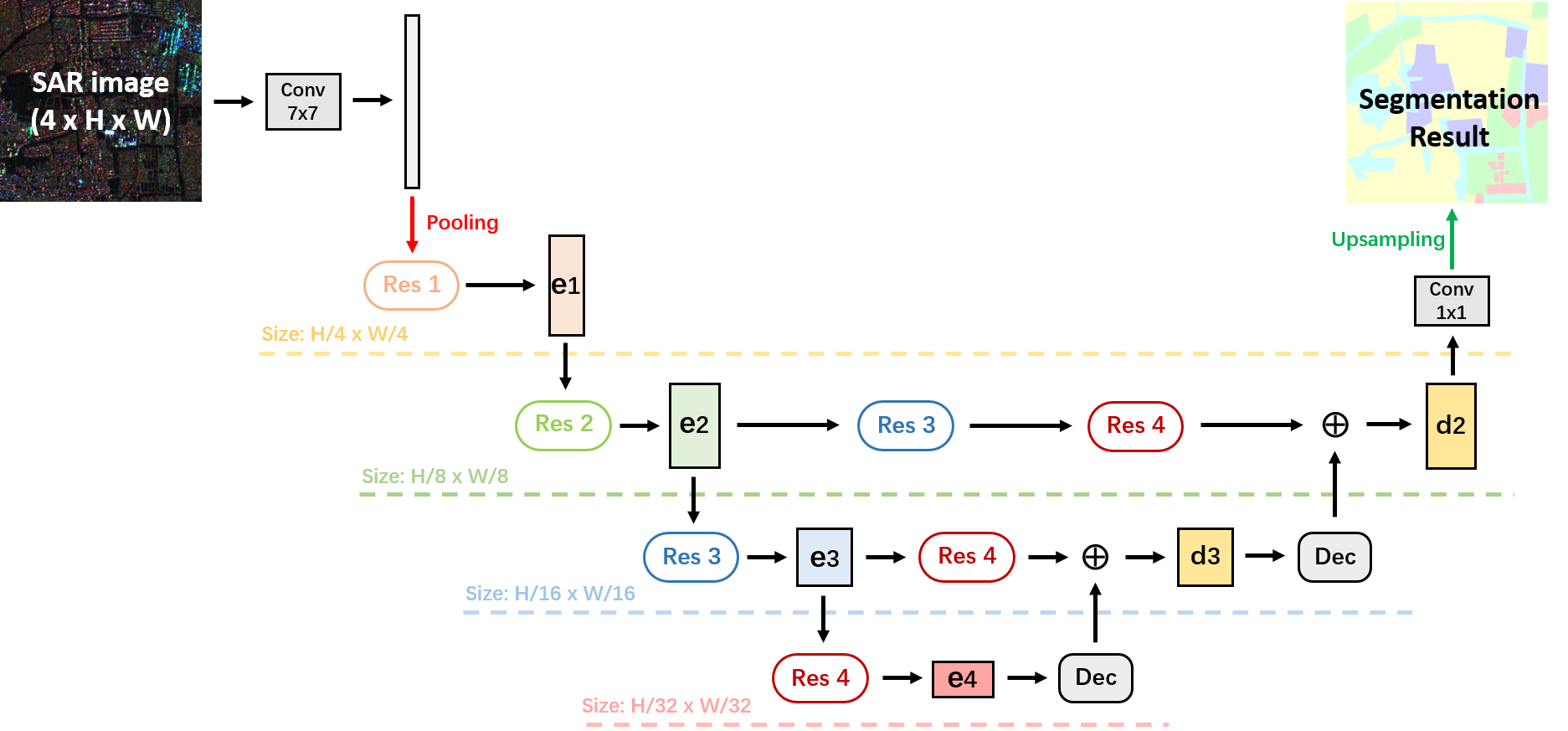}
    \caption{Atchitecture of the proposed Multi-path Residual Network (MP-ResNet).}
    \label{FigOverview}
\end{figure*}

\begin{figure}[htbp]
    \centering
    \includegraphics[width=8cm]{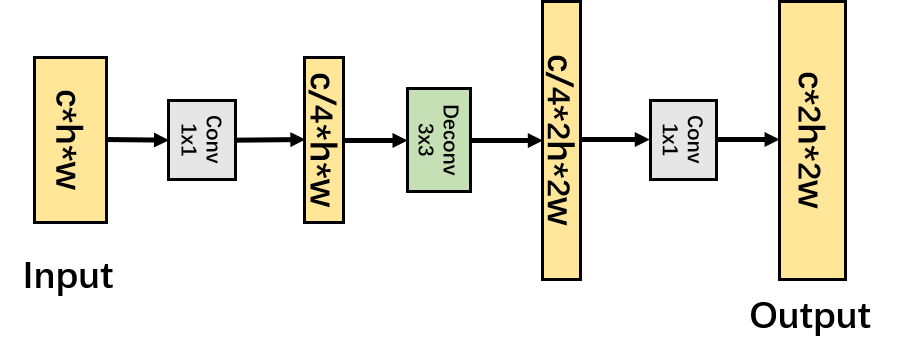}
    \caption{Feature deconvolution block of the proposed MP-ResNet.}
    \label{FigDeconv}
\end{figure}

Recently, with the emergence of several benchmark datasets, new developments on the CNN-based semantic segmentation of SAR images have been proposed. In \cite{mohammadimanesh2019new}, a 'encoder-decoder' CNN network with inception modules and skip connections is introduced for the semantic segmentation of wetland PolSAR images. In \cite{yue2020MANet}, a multi-scale attention based FCN (MANet) is presented combining multi-scale feature extraction and the attention mechanism. In \cite{wang2020hrsarnet}, a small yet efficient network (HR-SARNet) is proposed for the semantic segmentation of high-resolution SAR images. However, most of these literature studies are based on shallow CNNs to avoid over-fitting problems. Since shallow CNNs are not powerful enough to extract high-level semantic information, their accuracy is limited.
Differently from previous studies, this work aims to improve the semantic segmentation of high-resolution PolSAR images by proposing a novel network architecture, the Multi-path Residual Network (MP-ResNet). This network enables a multi-scale modeling of high-level semantic features through its parallel branches, which strengthens the learning of local discriminative features and reduces the effects of speckle noises. This network is based on a large benchmark dataset on PolSAR images made available by the Gaofen committee. Ablation studies show that the proposed approach achieves an large increase of 0.36\% and 0.64\% in terms of average OA and fwIoU, respectively, compared to the baseline FCN. It also surpasses several literature works in the comparative experiments.

\section{Methodology}

The key to improve the semantic segmentation of PolSAR images is to learn discriminative features from a larger image context, so that the effects of speckle noises can be mitigated. Based on this idea, we propose the Multi-path Residual Network (MP-ResNet) shown in Fig.\ref{FigOverview}. This network exploits a baseline fully convolutional network (FCN). It is inspired by the multi-path convolution design in high-resolution network (HRNet) \cite{wang2020hrnet} and the multi-scale feature fusion design in LinkNet \cite{chaurasia2017linknet}. In this section we introduce in details the design of the proposed MP-ResNet.

\subsection{Quick Embedding of Shallow Features}
PolSAR data generally contain 4 channels which are related to the 4 combination of linear polarizations (i.e. the HH and VV co-polarized channels and the HV and VH cross-polarized channels). The corresponding 4 images are stacked and given as input to a CNN. Since CNNs are capable of extracting semantic features from raw input data, no extra filtering operations have been applied to the input images. Instead, we merely apply maximum-suppression and normalization operations to stabilize and squeeze the value range of input data.

It is well-known that the signal at pixel level in PolSAR images is generally very noisy. The most discriminative features in these images are the signal values and texture patterns in local areas.  This makes it necessary to down-sample the features in early layers of a CNN and embed the features from a wider range. Considering this, we adopt the ResNet as the backbone feature extraction network. The first layers of ResNet are strided convolutions followed by a pooling layer, which quickly decreases the scaling rate of features to 1/4. Additionally, while 3$\times$3 convolutional kernels are widely used in CNNs to avoid over-fitting problems, the kernel size of the first convolutional layer in the ResNet is 7$\times$7. This helps the network to aggregate features from a larger pixel neighbourhood and alleviate the effect of speckle noises. After the first two residual modules of ResNet, the scaling rate of features is quickly decreased to 1/8. Considering that most of PolSAR images are not high-resolution and thus do not contain rich spatial details, this feature size is large enough for modeling the regional land-cover information. Therefore, 1/8 is set as the fundamental scaling rate for the aggregation of context information in the proposed network.

\subsection{Multi-path Semantic Information Embedding}
The size of valid receptive fields (VRFs) is known to be crucial to the embedding of context information in CNNs \cite{zhang2018context}. In the semantic segmentation of remote sensing data, the size of VRFs determines the spatial range from which the CNN can exploit discriminative features, which is related to the granularity of semantic segmentation results \cite{ding2020twostage}. Although the serial connection of pooling and strided convolutions can greatly enlarge the VRFs, it also brings the problem of losing spatial information. Therefore, how to simultaneously enlarge VRFs and preserve spatial information is one of the most crucial bottleneck problems in semantic segmentation tasks. PSPNet \cite{zhao2017pspnet} and DeepLab \cite{chen2018deeplabv3+} managed to enlarge VRFs without severe loss of spatial information through the use of additional context aggregation modules in the late layers of the CNNs. However, the context information is aggregated through pooling and dilated convolutions in these designs, which are less effective than stacked convolutional layers. The dilated convolutions may also cause gridding effects and enlarge computational costs.

Alternatively, HRNet presents a multi-path architecture that organizes multi-scale convolutional layers in a parallel manner. The highest scaling rate of HRNet is 1/32, which ensures a large VRF of the network. However, the serial convolutions in its parallel branches greatly increases the computational costs of this network. They may also cause over-fitting problems on small datasets. In addition, in HRNet the multi-scale feature branches are concatenated together to generate the semantic segmentation results, thus the features are not fully fused and utilized.

Inspired by the parallel feature embedding design of the HRNet, The proposed MP-ResNet uses ResNet as the feature extraction network but has 2 additional encoding branches after the second convolutional block. Therefore, the features are encoded both forwardly (size remain the same) and downstream-wise (size reduced). The same residual blocks are duplicated into the parallel branches so that each encoding branch has the same amount of convolutional layers. In this way, each encoding branch contains rich semantic information.

Differently from the HRNet, the parallel  branches in MP-ResNet focus on the embedding of high-level features (the lowest scaling rate is 1/8). Although the parallel embedding of larger feature maps are potentially feasible in other applications, for the semantic segmentation of PolSAR images our objective is to exploit discriminative features from a wider range. In this way, the segmentation results become less sensitive to pixel noise. The computational costs of MP-ResNet is only slightly higher than ResNet and far less than the HRNet (see discussion in Section \ref{sec.results}). Another significant difference of the MP-ResNet is that a decoder network is employed to fuse the features learned from its parallel branches.

\subsection{Fusion of Multi-scale Features}
Decoder networks are commonly used in semantic segmentation to recover the spatial details of encoded features. A common design is to concatenate or add the encoded features with the features from early layers of the encoder networks (e.g. UNet\cite{ronneberger2015unet} and SegNet\cite{badrinarayanan2017segnet}). Although this design can aggregate spatial information from low-level features, it also introduces 
redundant information (minor details and noises). For the segmentation of high-resolution PolSAR images this problem can be critical. However, in the proposed MP-ResNet, the fused multi-scale features are the ones encoded by the parallel branches of the encoder. These features contain rich semantic information, thus their fusion does not lead to noise problems. A feature deconvolution module \cite{chaurasia2017linknet} is introduced to enlarge the spatial size of features from higher branches before the fusion. Fig.\ref{FigDeconv} shows this spatial deconvolution block. It adopts a channel-wise 'Bottleneck' design to reduce the computation and refine the crucial semantic information. In this way, the multi-branch features are fused in a coarse-to-fine manner in the decoder.

\section{Experimental Results}\label{sec.results}
\subsection{Dataset Descriptions and Evaluation Metrics}
The experiments dataset of this study are developed on the Gaofen dataset provided by the '2020 Gaofen contest on automated high-resolution earth observation image interpretation'. The PolSAR images are collected from the Gaofen-3 satellite. Their ground sampling distance is between 1m and 3m. The ground truth maps are annotated according to 5 land-cover types: background, built-up area, vegetation, water and bare soil. The accessible training data are 500 pairs of PolSAR images and label maps each with 512$\times$512 pixels. The testing data are not visible to users, but a scoring system is provided to evaluate the uploaded algorithms.

We adopts 3 metrics for the evaluation of semantic segmentation results. They are overall accuracy (OA), F1 score and frequency weighted intersection over union (fwIoU). OA is the ratio of the number of correctly classified pixels among all pixel numbers. F1 is calculated as:
    \begin{equation}
    \rm F1=2 \cdot {\frac{precision \cdot recall}{precision+recall}}
    \label{F1}
    \end{equation}
FwIoU is the evaluation metric suggested by the contest organizer. It is calculated as:
    \begin{equation}
        fwIoU = \frac{1}{\sum_{i=0}^{N}\sum_{j=0}^{N}}\sum_{i=0}^{N}\frac{\sum_{j=0}^{N}s_{ij}s_{ii}}{\sum_{j=0}^{N}s_{ij}+\sum_{j=0}^{N}s_{ji}-s_{ii}}
    \end{equation}
where N is the number of total classes, $s_{ij}$ denotes the number of i-th class pixels that are classified into the j-th class.

\subsection{Multi-fold Ablation Study}\label{sec.ablation}
To quantitatively evaluate the improvement of the proposed MP-ResNet over the baseline method (FCN), we conduct a multi-fold ablation study on the Gaofen dataset. The training and validation sets are randomly divided from all available training data (500 image pairs) with a numeric ratio of 9:1. In this way, a total of 10 training and validation sets are obtained. To reduce the effects of random factors, the ablation study has been conducted on all the 10 training-validation sets. The results are reported in table \ref{Table.Ablation}. Compared to the baseline method (FCN), the proposed MP-ResNet shows average improvements of 0.36\%, 0.54\% and 0.64\% in OA, mean F1 and fwIoU, respectively. The improvements have also been verified on the contest test set which is not directly available. According to the fwIoU scores obtained from the contest, the proposed MP-ResNet has an increase of 1.23\% in fwIoU compared to the FCN.

\begin{table}[htbp]
    \centering
    \caption{Results of the multi-fold ablation study.}
    \resizebox{1\linewidth}{!}{%
        \begin{tabular}{c|ccc|ccc}
        \toprule
        \multirow{2}*{Datasets} & \multicolumn{3}{c|}{FCN (Baseline)} & \multicolumn{3}{c}{MP-ResNet (Proposed)}\\
        \cline{2-7}
         & OA(\%) & mF1(\%) & fwIoU(\%) & OA(\%) & mF1(\%) & fwIoU(\%)\\
        \hline
         Val 1 & 93.08 & 91.33 & 88.14 & 93.95(\textcolor{green}{+0.87}) & 92.47(\textcolor{green}{+1.14}) & 89.63(\textcolor{green}{+1.49}) \\
         Val 2 & 93.41 & 91.81 & 89.04 & 93.88(\textcolor{green}{+0.47}) & 92.58(\textcolor{green}{+0.77}) & 89.88(\textcolor{green}{+0.84}) \\
         Val 3 & 91.70 & 88.20 & 86.35 & 91.86(\textcolor{green}{+0.16}) & 88.24(\textcolor{green}{+0.04}) & 86.71(\textcolor{green}{+0.36}) \\
         Val 4 & 90.08 & 85.52 & 84.65 & 90.22(\textcolor{green}{+0.14}) & 86.11(\textcolor{green}{+0.59}) & 84.58(\textcolor{green}{+0.24}) \\
         Val 5 & 93.06 & 89.64 & 88.34 & 93.45(\textcolor{green}{+0.39}) & 90.46(\textcolor{green}{+0.82}) & 88.97(\textcolor{green}{+0.63}) \\
         Val 6 & 91.96 & 89.28 & 86.37 & 92.85(\textcolor{green}{+0.89}) & 90.76(\textcolor{green}{+1.48}) & 87.81(\textcolor{green}{+1.44}) \\
         Val 7 & 92.97 & 89.22 & 88.25 & 93.33(\textcolor{green}{+0.36}) & 89.20(\textcolor{red}{-0.02}) & 88.90(\textcolor{green}{+0.65}) \\
         Val 8 & 91.43 & 87.21 & 86.09 & 91.60(\textcolor{green}{+0.17}) & 87.58(\textcolor{green}{+0.37}) & 86.52(\textcolor{green}{+0.43}) \\
         Val 9 & 93.11 & 90.66 & 88.66 & 93.24(\textcolor{green}{+0.13}) & 90.72(\textcolor{green}{+0.06}) & 88.91(\textcolor{green}{+0.25}) \\
         Val 10 & 93.40 & 91.59 & 89.50 & 93.62(\textcolor{green}{+0.22}) & 91.76(\textcolor{green}{+0.17}) & 89.85(\textcolor{green}{+0.35}) \\
         Test set & - & - & 69.42 & - & - & 70.65 (\textcolor{green}{+1.23}) \\
        \hline
         \multicolumn{4}{c|}{Average Improvements} & +0.36 & +0.54 & +0.64 \\
        
        \hline
        \end{tabular}}
   \label{Table.Ablation}
\end{table}

\subsection{Comparative Experiments}
To further assess the improvements brought by the proposed MP-ResNet, we compared it with several literature works. Apart from the baseline method FCN, several well-established methods in the field of semantic segmentation have been considered, including the SegNet \cite{badrinarayanan2017segnet}, the UNet \cite{ronneberger2015unet}, the PSPNet \cite{zhao2017pspnet} and the DeepLabv3+ \cite{chen2018deeplabv3+}. Since the proposed MP-ResNet is inspired by the multi-path architecture of HRNet\cite{wang2020hrnet} and the feature fusion design of LinkNet\cite{chaurasia2017linknet}, we also included these networks in comparison. Moreover, several literature methods presented for the semantic segmentation of SAR images have also been tested, including the mult-scale FCN (MS-FCN) in \cite{wu2019polsar}, the Inception FCN (Inc-FCN) in \cite{mohammadimanesh2019new} and the HR-SARNet \cite{wang2020hrsarnet}. The training and validation sets used in the comparison are these related to the first row of table \ref{Table.Ablation}. For fairness, the same parameter settings have been applied during the training process (e.g. training epochs, batch size, learning rate).

Table \ref{Table.Compare} presents the quantitative results obtained on the Gaofen dataset. Due to the effects of speckle noise, the performances of shallow networks (SegNet, HR-SARNet, Inc-FCN and UNet), which rely heavily on low-level features, is unsatisfactory. Although there is also a multi-scale design in the MS-FCN, there is no enhancement in its branches, thus its accuracy is lower than that of the FCN. The simple FCN without any sophisticated design ranks at the 3rd place. Although HRNet shows better performance in the semantic segmentation of VHR optical remote sensing images \cite{zhang2020multi}, its size is too large to be fully trained on the Gaofen dataset, thus its accuracy is far lower than that of the FCN. LinkNet also has a deep encoding network (ResNet34) but it has skip connections with low-level features, which introduce noise and degrades its accuracy. The PSPNet has an additional multi-scale average pooling head compared with the FCN, which does not prove to be effective on the PolSAR dataset. The multi-scale atrous convolution head in DeepLabV3+ increases the mean F1, OA and fwIoU by 0.57\%, 0.30\% and 0.36\%, respectively, compared to the FCN. The proposed MP-ResNet achieves the best accuracy in nearly every metrics except the F1 of the 'others' class. Its improvements over the DeepLabV3+ are 0.57\%, 0.57\% and 1.13\% in mean F1, OA and fwIoU, respectively. This proves the effectiveness of the proposed network with which we have won the 2nd place in the 'Gaofen Challenge' contest.

\begin{table*}[htbp]
    \centering
    \caption{Quantitative Results of the comparative study.}
    \resizebox{1\linewidth}{!}{%
        \begin{tabular}{c|cccccc|ccc}
        \toprule
        \multirow{2}*{Methods} & \multicolumn{6}{c|}{F1(\%)} & \multirow{2}*{mean F1(\%)} & \multirow{2}*{OA(\%)} & \multirow{2}*{fwIoU(\%)$\blacktriangledown$} \\
        \cline{2-7}
         & Water & Built-up area & Industrial area & Grassland & Barren & Others & & & \\
        \hline
        SegNet\cite{badrinarayanan2017segnet} & 45.78 & 24.13 & 36.42 & 65.80 & 0.00 & 0.00 & 28.69 & 45.79 & 36.25\\
        HR-SARNet\cite{wang2020hrsarnet} & 73.10 & 76.29 & 57.31 & 78.67 & 0.00 & 0.10 & 47.58 & 74.01 & 62.87\\
        Inc-FCN\cite{mohammadimanesh2019new} & 83.62 & 74.90 & 52.92 & 78.58 & 4.41 & 15.05 & 51.58 & 73.74 & 62.97\\
        MS-FCN\cite{wu2019polsar} & 69.68 & 77.14 & 60.02 & 79.13 & 0.00 & 0.01 & 47.66 & 74.63 & 63.09\\
        UNet\cite{ronneberger2015unet} & 83.58 & 77.54 & 59.63 & 79.67 & 0.02 & 22.36 & 53.80 & 76.14 & 65.43\\
        HRNet\cite{wang2020hrnet} & 91.30 & 90.19 & 85.09 & 90.34 & 73.96 & 92.98 & 87.31 & 90.02 & 83.41\\
        LinkNet\cite{chaurasia2017linknet} & 90.18 & 91.63 & 88.43 & 91.11 & 74.95 & 92.90 & 88.20 & 91.17 & 85.19\\
        PSPNet\cite{zhao2017pspnet} & 92.71 & 93.36 & 89.71 & 93.12 & 84.03 & 94.91 & 91.31 & 92.98 & 88.05\\
        FCN\cite{long2015fcn} & 91.89 & 93.53 & 90.26 & 93.20 & 84.79 & 94.28 & 91.33 & 93.08 & 88.14\\
        DeepLabV3+\cite{chen2018deeplabv3+} & 93.44 & 93.67 & 90.37 & 93.41 & 85.15 & \textbf{95.32} & 91.90 & 93.38 & 88.50\\
        MP-ResNet (proposed) & \textbf{94.85} & \textbf{94.29} & \textbf{90.78} & \textbf{93.93} & \textbf{86.03} & 94.96 & \textbf{92.47} & \textbf{93.95} & \textbf{89.63}\\
        \hline
        \end{tabular}}
   \label{Table.Compare}
\end{table*}

\begin{figure}[thpb]
\centering
    {\includegraphics[height=0.5cm]{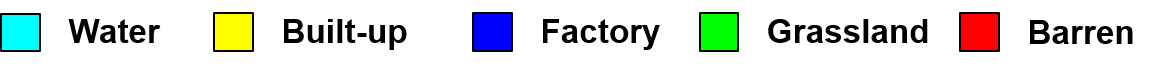}}\\
    \setlength{\tabcolsep}{1.0pt}
    \begin{tabular}{>{\centering\arraybackslash}m{1.7cm}>{\centering\arraybackslash}m{1.7cm}>{\centering\arraybackslash}m{1.7cm}>{\centering\arraybackslash}m{1.7cm}>{\centering\arraybackslash}m{1.7cm}}
        \includegraphics[width=1.7cm]{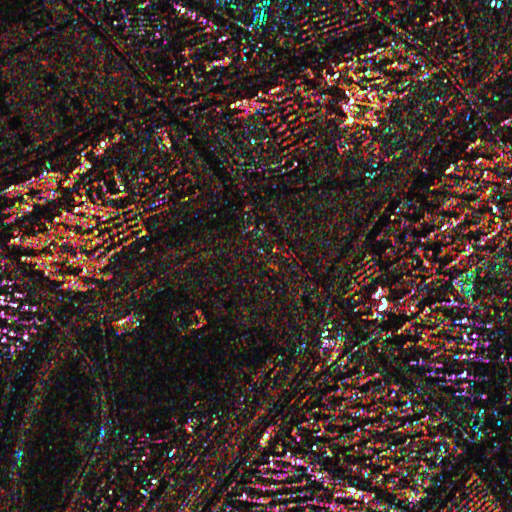} &
        \includegraphics[width=1.7cm]{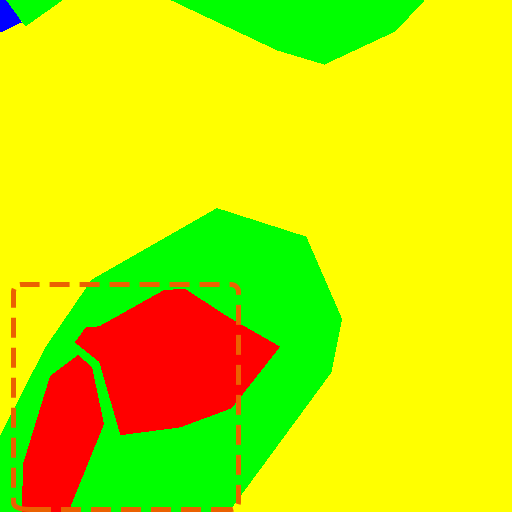} &
        \includegraphics[width=1.7cm]{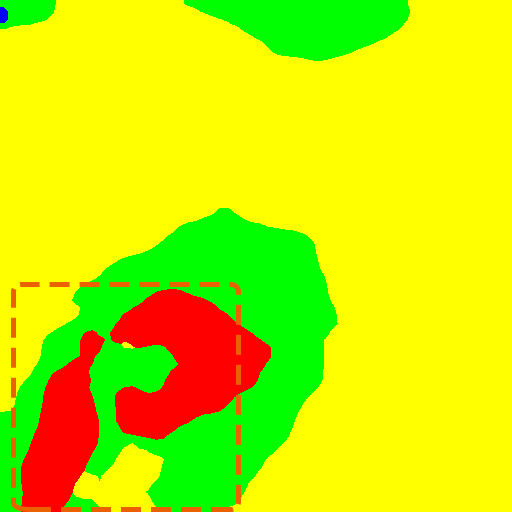} &
        \includegraphics[width=1.7cm]{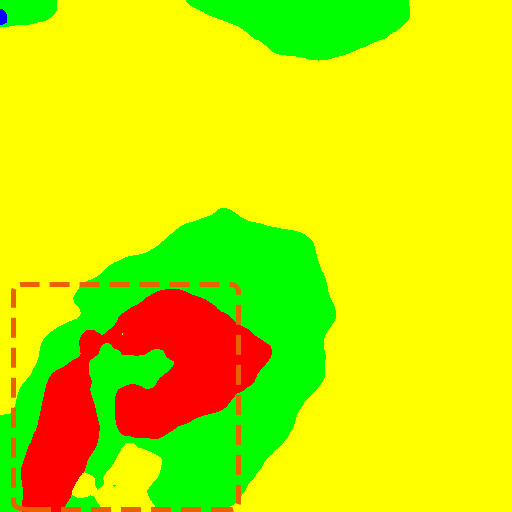} &
        \includegraphics[width=1.7cm]{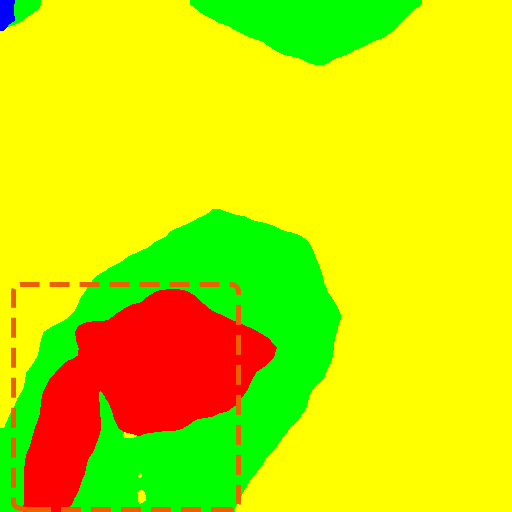}\\
        \includegraphics[width=1.7cm]{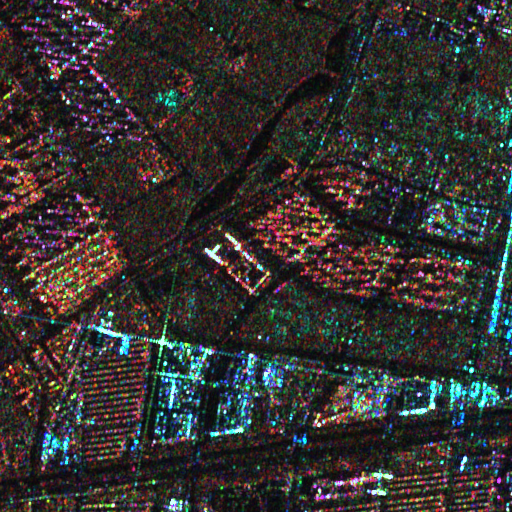} &
        \includegraphics[width=1.7cm]{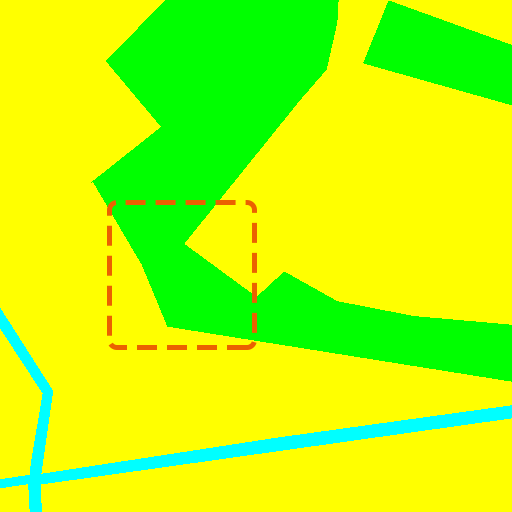} &
        \includegraphics[width=1.7cm]{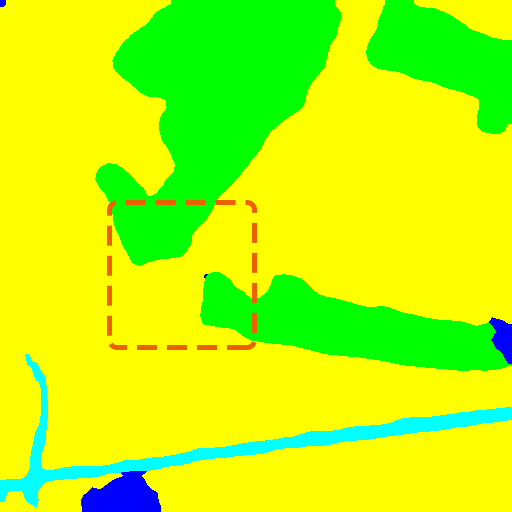} &
        \includegraphics[width=1.7cm]{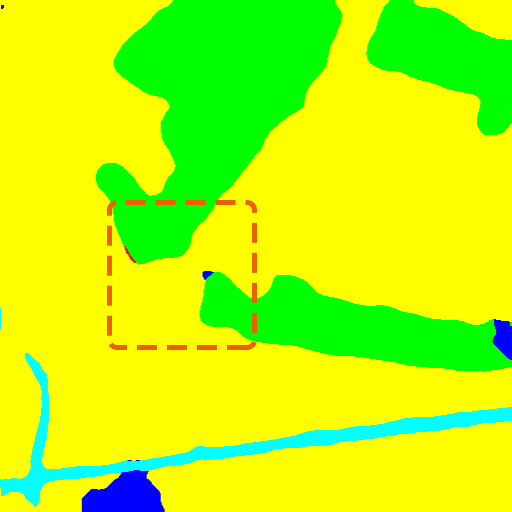} &
        \includegraphics[width=1.7cm]{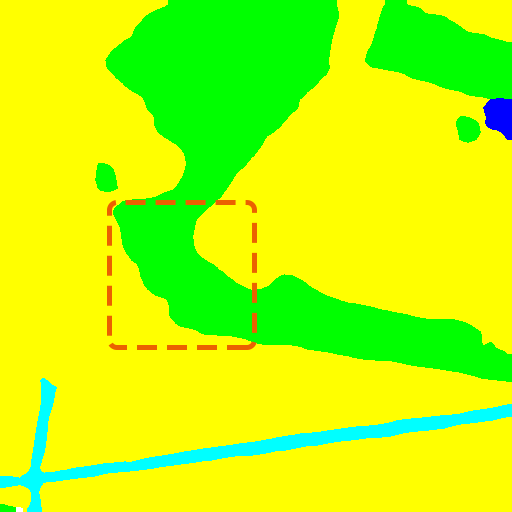}\\
        \includegraphics[width=1.7cm]{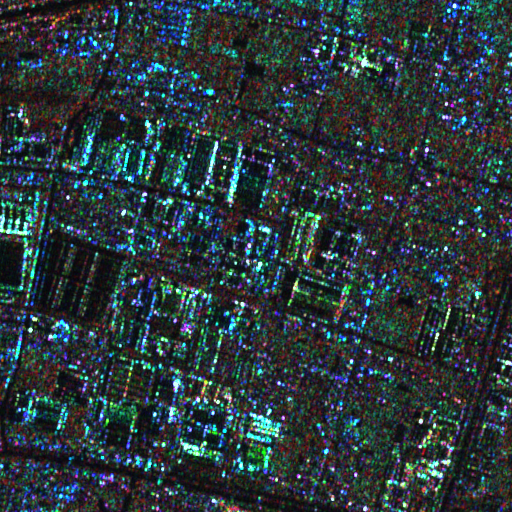} &
        \includegraphics[width=1.7cm]{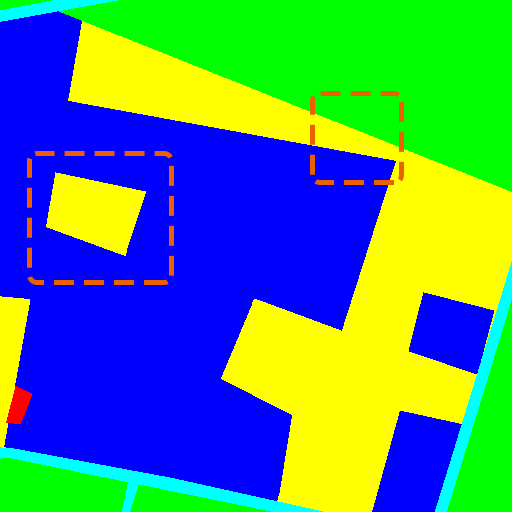} &
        \includegraphics[width=1.7cm]{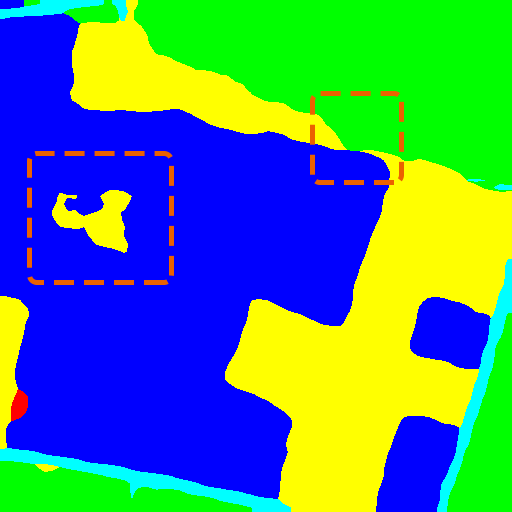} &
        \includegraphics[width=1.7cm]{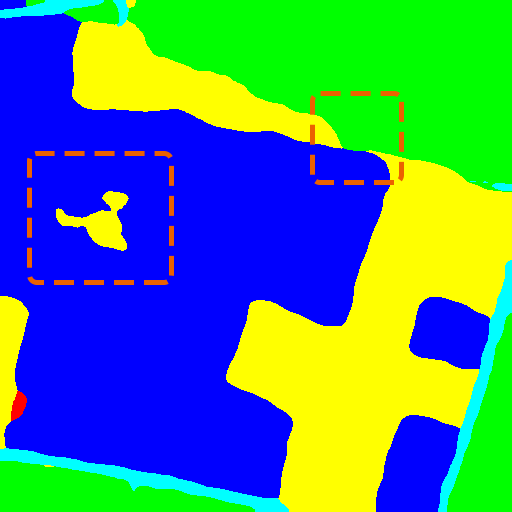} &
        \includegraphics[width=1.7cm]{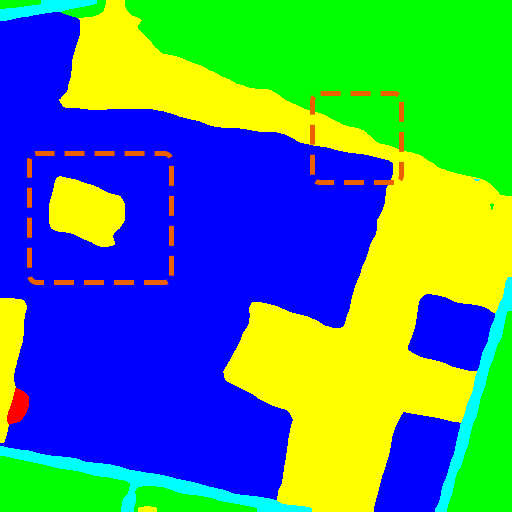}\\
        \includegraphics[width=1.7cm]{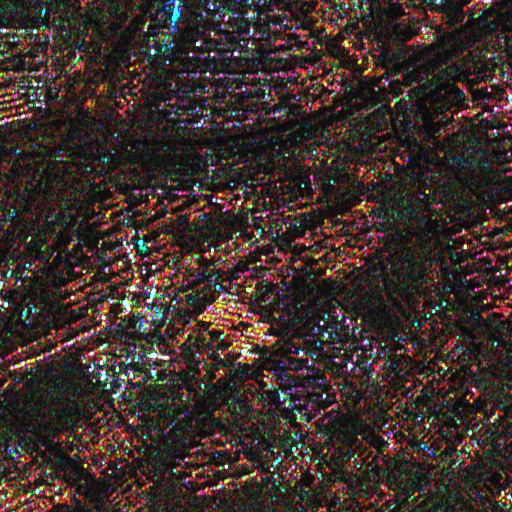} &
        \includegraphics[width=1.7cm]{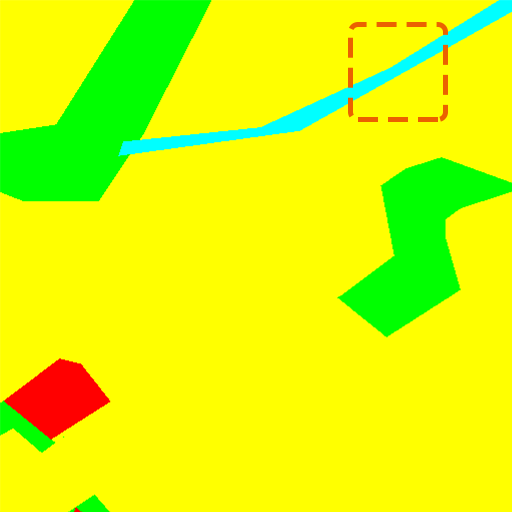} &
        \includegraphics[width=1.7cm]{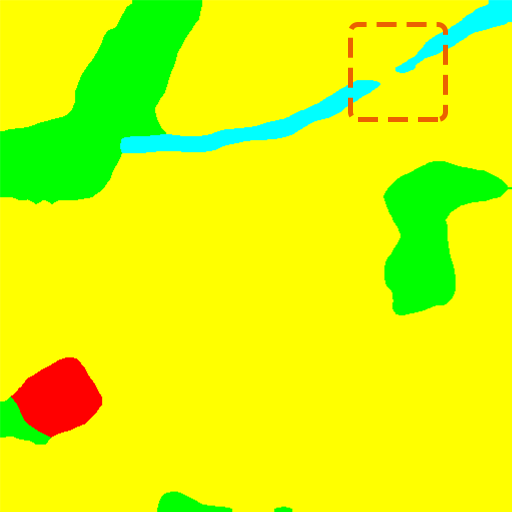} &
        \includegraphics[width=1.7cm]{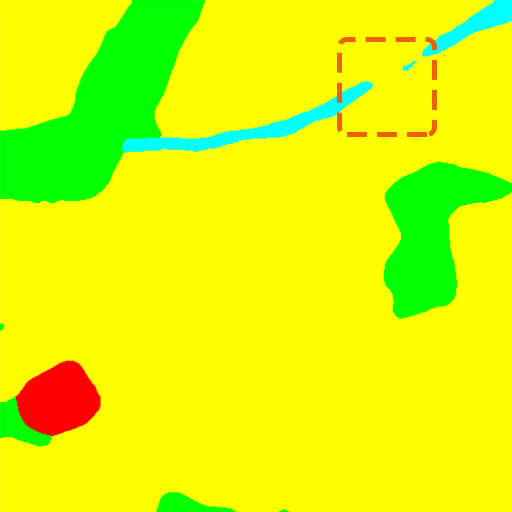} &
        \includegraphics[width=1.7cm]{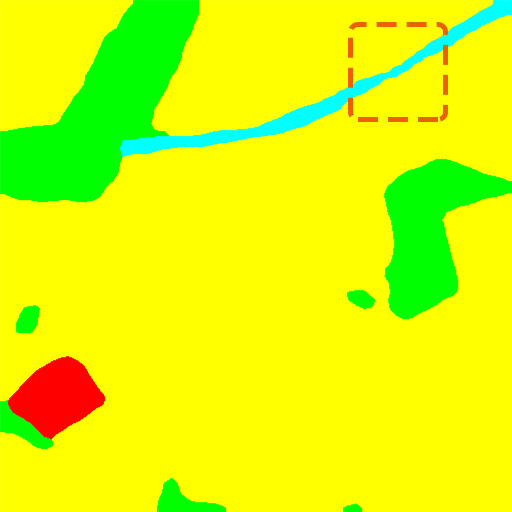}\\
        PolSAR images (pseudo color) & Ground truth & FCN & Deeplabv3+ & MP-ResNet (Proposed)\\
    \end{tabular}
    \caption{Comparison of segmented maps obtained by different methods on sample testing areas.}
    \label{Fig.Compare}
\end{figure}

Fig.\ref{Fig.Compare} shows a comparison of the segmentation results on several sample areas. Due to its multi-path modeling and multi-scale feature fusion design, the proposed MP-ResNet is capable of modeling context information from a wider image range. Therefore, some critical areas for other networks are correctly segmented and the detected object boundaries are more continuous. 

\begin{table}[htbp]
\centering
    \caption{Comparison of model size and computational cost expressed in terms of params (Mb) and FLOPS (Gbps), respectively.}
    \begin{tabular}{l|cc}
        \toprule
        Methods & Params (Mb) & FLOPS (Gbps) $\blacktriangledown$\\
        \hline
        HR-SARNet\cite{wang2020hrsarnet} & 0.06 & 6.12 \\
        MS-FCN\cite{wu2019polsar} & 23.54 & 21.73 \\
        LinkNet \cite{chaurasia2017linknet} & 21.65 & 27.76\\
        FCN \cite{long2015fcn} & 21.35 & 90.97 \\
        HRNet\cite{wang2020hrnet} & 65.85 & 93.64 \\
        DeeplabV3+ \cite{chen2018deeplabv3+} & 22.29 & 94.84  \\
        PSPNet\cite{zhao2017pspnet} & 22.73 & 95.55\\
        MP-ResNet (proposed) & 54.97 & 115.93 \\
        UNet \cite{ronneberger2015unet} & 9.16 & 221.68 \\
        Inc-FCN\cite{mohammadimanesh2019new} & 6.14 & 329.37\\
        SegNet\cite{badrinarayanan2017segnet} & 3.51 & 341.06 \\
        \bottomrule
    \end{tabular}
    \label{Table.Calculations}
\end{table}

Table \ref{Table.Calculations} presents the sizes and computational costs of the compared methods. The floating point operations per second (FLOPS) are calculated based on the input size [4, 512, 512] of the images in the Gaofen dataset. The literature methods for the semantic segmentation of SAR images (HR-SARNet and Inc-FCN) are generally shallow, thus and their sizes are relatively small. UNet, Inc-FCN and SegNet need the most FLOPS since they apply many convolution operations on the early-layer features. Although the parameter size of the MP-ResNet is much larger than the FCN, its FLOPS do not increase significantly.

\section{Conclusion}
The semantic segmentation of PolSAR images is challenging due to the intense speckle noise and the lack of large training datasets. Taking advantage of the open dataset from the Gaofen contest, we propose a Multi-Path Residual Network (MP-ResNet) for the semantic segmentation of high-resolution PolSAR images. Compared to the baseline FCN, the MP-ResNet has three parallel semantic embedding branches to strengthen the aggregation of context information. It also adopts a multi-scale feature fusion design in its decoder to take advantage from each encoding branch. As a result, the VRF of the MP-ResNet is significantly enlarged, thus allowing the aggregation of discriminative features from a large range and alleviating the impact of noise. The multi-fold ablation study conducted on the Gaofen dataset has proved the effectiveness of our designs. The comparative experiments with several state-of-the-art methods show that the proposed method has a significant improvement in all accuracy metrics.

Since the objectiveness of this work is to propose a general architecture for the semantic segmentation of PolSAR images, we did not add sophisticated designs. However, in future works we plan to combine additional context aggregation modules to improve the performance (e.g., dilated convolutions, attention mechanisms).



\ifCLASSOPTIONcaptionsoff
  \newpage
\fi

\bibliographystyle{IEEEtran}
\bibliography{refs}

\end{document}